\documentclass[twocolumn]{extarticle}
\usepackage[latin9]{inputenc}
\usepackage{array}
\usepackage{booktabs}
\usepackage{url}
\usepackage{multirow}
\usepackage{graphicx}
\usepackage[unicode=true,
 bookmarks=false,
 breaklinks=false,pdfborder={0 0 1},backref=section,colorlinks=false]
 {hyperref}
\usepackage{authblk}

\makeatletter

\providecommand{\tabularnewline}{\\}

\@ifundefined{date}{}{\date{}}

\usepackage{microtype}
\usepackage{colortbl}
\usepackage{xcolor}

\usepackage{xspace}
\newcommand*{\eg}{e.g.\@\xspace}
\newcommand*{\ie}{i.e.\@\xspace}

\@ifundefined{showcaptionsetup}{}{%
 \PassOptionsToPackage{caption=false}{subfig}}
\usepackage{subfig}
\makeatother

\begin{document}

\title{Intriguing Properties of Randomly Weighted Networks: Generalizing
While Learning Next to Nothing}

\author[ ]{Amir Rosenfeld}
\author[ ]{John K.Tsotsos}
\affil[ ]{Department of Electrical Engineering and Computer Science\protect\\York University
\protect\\Toronto, ON, Canada, M3J 1P3}
\affil[ ]{\texttt {\{amir@eecs,tsotsos@cse\}.yorku.ca}}

\maketitle
\begin{abstract}
Training deep neural networks results in strong learned representations
that show good generalization capabilities. In most cases, training
involves iterative modification of all weights inside the network
via back-propagation. In Extreme Learning Machines, it has been suggested
to set the first layer of a network to fixed random values instead
of learning it. In this paper, we propose to take this approach a
step further and fix almost all layers of a deep convolutional neural
network, allowing only a small portion of the weights to be learned.
As our experiments show, fixing even the majority of the parameters
of the network often results in performance which is on par with the
performance of learning all of them. The implications of this intriguing
property of deep neural networks are discussed and we suggest ways
to harness it to create more robust representations. 
\end{abstract}

\section{Introduction\label{intro}}

Deep neural networks create powerful representations by successively
transforming their inputs via multiple layers of computation. Much
of their expressive power is attributed to their depth; theory shows
that the complexity of the computed function grows exponentially with
the depth of the net \cite{raghu2016expressive}. This renders deep
networks more \emph{expressive} than their shallower counterparts
with the same number of parameters. Moreover, the data representation
is more efficient from an information-theoretic point of view \cite{shwartz2017opening}.
This has led to increasingly deeper network designs, some over a thousand
layers deep \cite{he2016identity}.

Modern day architectures \cite{krizhevsky2012imagenet,simonyan2014very,he2016identity,journals/corr/ZagoruykoK16}
contain millions to billions of parameters \cite{shazeer2017outrageously}
- often exceeding the number of training samples (typically ranging
from tens of thousands \cite{krizhevsky2009learning} to millions
\cite{russakovsky2015imagenet}). This suggests that these networks
could be prone to over-fitting, or are otherwise highly-overparameterized
and could be much more compact; this is supported by network pruning
methods, such as \cite{li2016pruning}, which are able to retain network
accuracy after removing many of the weights and re-training. Counter-intuitively,
\cite{1801.10447} have even shown that a network can be pruned by
selecting an arbitrary subset of filters and still recover the original
accuracy, hinting at a large redundancy in the parameter space. The
large parameter space may explain why current methods in machine learning
tend to be so data-hungry. Could be it that not all of the weights
require updating, or are equally useful (this is suggested by \cite{raghu2016expressive})? 

The common optimization pipeline involves an iterative gradient-based
method (\eg SGD), used to update all weights of the network to minimize
some target loss function. Instead of training all weights, we suggest
an almost extreme opposite: network weights are initialized randomly
and only a certain fraction is updated by the optimization process.
As our experiments show, while this does have a negative effect on
network performance, its magnitude is surprisingly small with respect
to the number of parameters not learned.

This effect holds for a range of architectures, conditions, and datasets,
ruling out the option that it is specific to a peculiar combination
thereof. We discuss and explore various ways of selecting subsets
of networks parameters to be learned. To the best of our knowledge,
while others have shown analytic properties of randomly weighted networks,
we are the first to explore the effects of keeping most of the weights
at their randomly initialized values in multiple layers. We claim
that successfully training mostly-random networks has implications
for the current understanding of deep learning, specifically: 
\begin{enumerate}
\item Popular network architectures are grossly over-parameterized
\item Current attempts at interpreting emergent representations inside neural
networks may be less meaningful than thought.
\end{enumerate}
Moreover, he ability to do so opens up interesting possibilities,
such as ``overloading'' networks by keeping a fixed backbone subset
of parameters and re-training only a small set. This can be used to
create cheap ensemble models which are nevertheless diverse enough
to outperform a single model. 

The rest of the paper is organized as follows. In Section \ref{sec:Related-Work}
we describe related work. This is followed by a description of our
method (Section \ref{sec:Method}), and an extensive set of experiments
to limit the learned set of parameters in various ways. We end with
some discussion and concluding remarks. For reproducibility, we will
make code publicly available.

\section{Related Work\label{sec:Related-Work}}

\paragraph{Random Features}

There is a long line of research revolving around the use of randomly
drawn features in machine learning. Extreme Learning Machines show
the utility of keeping some layer of a neural net fixed - but this
is usually done only for one or two layers, and not within layers
\cite{huang2015trends} or across multiple (more than two) layers.
\cite{rudi2017generalization} has shown how picking random features
has merits over matching kernels to the data. \cite{giryes2015deep}
have analytically shown useful properties of random nets with Gaussian
weights. As mentioned in the work of \cite{raghu2016expressive},
many of the theoretical works on deep neural networks assume specific
conditions which are not known to hold in practice; we show empirically
what happens when weights are selected randomly (and fixed) throughout
various layer of the network and within layers. 

\paragraph{Fixed Features }

A very recent result is that of \cite{hoffer2018fix}, showing - quite
surprisingly - that using a fixed, Hadamard matrix \cite{horadam2012hadamard}
for a final classification layer does not hinder the performance of
a classifier. In contrast, we do not impose any constraints on the
values of any of the fixed weights (except drawing them from the same
distribution as that of the learned ones), and evaluate the effect
of fixing many different subsets of weights throughout the network. 

\paragraph{Net Compression/Filter Pruning}

Many works attempt to learn a compact representation by pruning unimportant
filters: for example, compressing the network after learning \cite{li2016pruning,han2015deep,han2015learning,1801.10447};
performing tensor-decompositions on the filter representations \cite{liu2015sparse}
or regularizing their structure to have a sparse representation \cite{wen2016learning};
and designing networks which are compact to begin with, either by
architectural changes \cite{iandola2016squeezenet,howard2017mobilenets},
or by learning\textbf{ }discrete weights , \eg \cite{shayar2017learning,rastegari2016xnor}. 

\paragraph{Network Interpretability}

There are many works attempting to dissect the representation learned
within networks, in order to develop some intuition about their inner
workings, improve them by ``debugging'' the representation or extract
some meaningful explanation about the final output of the network.
The work of \cite{zeiler2014visualizing} analyzes feature selectivity
as network depth progresses. They also attempt to map the activities
of specific filters back into pixel-space. Other methods mapping the
output of the network to specific image regions has been suggested,
either using gradient-based methods \cite{zhou2016learning} or biologically
inspired ones \cite{biparva2017stnet,zhang2016top}. Others either
generate images that maximize the response of an image to a specific
category \cite{simonyan2013deep}) or attempt to invert internal representations
in order to recover the input image \cite{mahendran2015understanding}.
Another line of work, such as that of \cite{bau2017network} shows
how emergent representations relate to ``interpretable'' concepts.
Some have tried to supervise the nets to become more interpretable
\cite{dong2017towards}. The interpretability of a network is often
defined by some correlation between a unit within the net and some
concept with semantic meaning. As we keep most features random, we
argue that (nearly) similarly powerful features can emerge without
necessarily being interpretable. Admittedly, interpretability can
be defined by examining a population of neurons, \eg \cite{fong2018net2vec},
but this kind of interpretation depends on some function which learns
to map the outputs of a set of neurons to a given concept. In this
regard, any supervised network is by construction interpretable. 
\begin{figure*}
\begin{centering}
\includegraphics[width=1\textwidth]{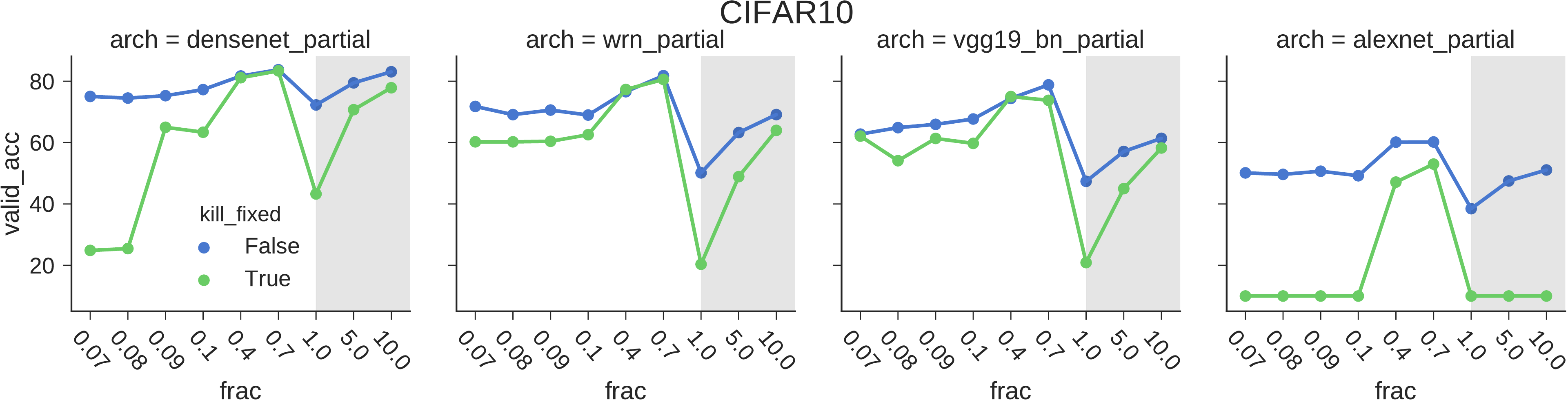}
\par\end{centering}
\includegraphics[width=1\textwidth]{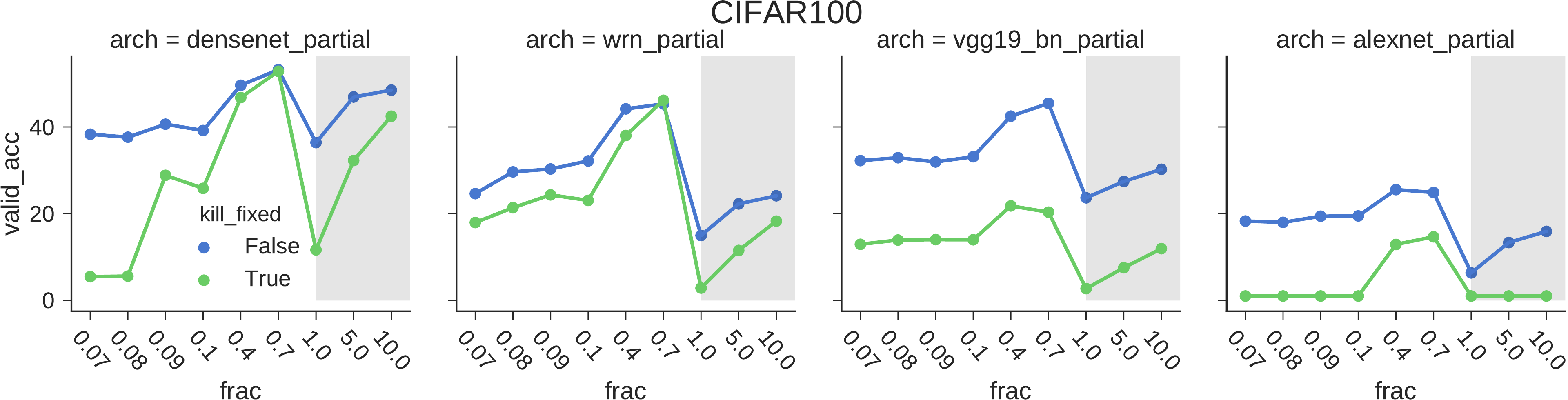}

\caption{\label{fig:Training-only-a}Training with only a few parameters (10
epochs): deep networks can generalize surprisingly well when only
a small fraction of their parameters is learned. Y-axis is validation
top-1\% accuracy. Unshaded area specifies fraction of filters learned
for at each convolutional layer. Shaded areas specify a fixed integer
number of filters learned at each conv. layer. Blue lines represent
training all except the specified frozen fraction. Green lines represent
all but specified fraction of weights zeroed out. }
\end{figure*}

\section{Method: Learning Partial Networks\label{sec:Method}}

In standard settings, all weights of a neural network $N$ are learned
in order to minimize some loss function $L$. Our goal is to test
how many parameters actually have to be learned - and what effect
fixing some of the parameters has on the final performance. 

In this work we concern ourselves with vision-related tasks dealing
with image classification, though in the future we intend to show
it on non-vision (e.g, language) tasks as well. In this setting, the
network is typically defined by a series of convolutional layers,
interleaved by non-linearities (e.g, ReLU). The final layer is usually
a fully-connected one, though it can be cast as a form of convolutional
layer as well. 

We proceed with some definitions. Let $\mathcal{W}$ be the set of
all parameters of a network $N$. Assume $N$ has $n$ layers. All
of our experiments can be framed as splitting $\mathcal{W}$ into
two disjoint sets: $\mathcal{W}=\mathcal{W}_{f}\cup\mathcal{W}_{l}$.
In each experiment we fix the weights of $\mathcal{W}_{f}$ and allow
$\mathcal{W}_{l}$ to be updated by the optimizer. $\mathcal{W}_{f}$
are either randomly initialized or set to zero. $W_{l}=\cup_{i\in1}^{n}w_{l,i}$
is a partition of $W_{l}$ into a selected set of weights $w_{l,i}$
for each layer $i\in\{1\dots n\}.$ Let $M_{i}\in\mathcal{R}^{O_{i}\times I_{i}\times k_{i}\times k_{i}}$
be the tensor defining the filters of layer $i$ of $N$, where: $O_{i}$
is the number of output channels (a.k.a number of filters), $I_{i}$
is the number of input channels, and $k_{i}$ is the kernel size. 

For each convolutional layer, the corresponding $w_{l,i}$ defines
a slice of some dimension of $M_{i}$:
\begin{itemize}
\item Slicing the first dimension this produces a subset of the filters,
which will be learned: $f_{i,j}\subseteq\mathcal{F}{}_{i}$ where
$\mathcal{F}{}_{i}$ is the set of filters of this layer. 
\item Slicing the second dimension, $w_{l,i}$ allows to learn only some
of the incoming connections for each filter.
\item Slicing the third and fourth dimensions $w_{l,i}$ allows to learn
only some of the spatial locations of each filter. 
\end{itemize}
We note that for layers with a bias term we do not split along any
dimension other than the first. In addition, we keep fully-connected
layers intact - we either learn them entirely or not at all. 

For simplicity, we treat all filters in a homogeneous manner, \eg,
no set of filters, such as ``shortcut'' filters used in resnets
\cite{journals/corr/ZagoruykoK16} is treated in a special way. In
addition, selecting a subset of coefficients in some dimension is
always implemented by choosing the first $p$ coefficients, where
$p$ and the dimension depend on the specifics of the scenario we
are currently testing. Admittedly, this could lead to suboptimal results
(see below). However, the goal here is not to learn an optimal sparse
set of connections (cf. \cite{liu2015sparse,wen2016learning}, but
rather to show the power inherent in choosing an arbitrary subset
of a given size. 

Network training proceeds as usual, via back-propagation - only that
the elements of $\mathcal{W}_{f}$ are treated as constants. We use
Momentum-SGD with weight-decay for the optimization. Next, we describe
various ways in to test how well the network converges to a solution
for different configurations of $w_{l}$. 
\begin{figure*}
\begin{centering}
\subfloat[]{\includegraphics[width=0.8\columnwidth]{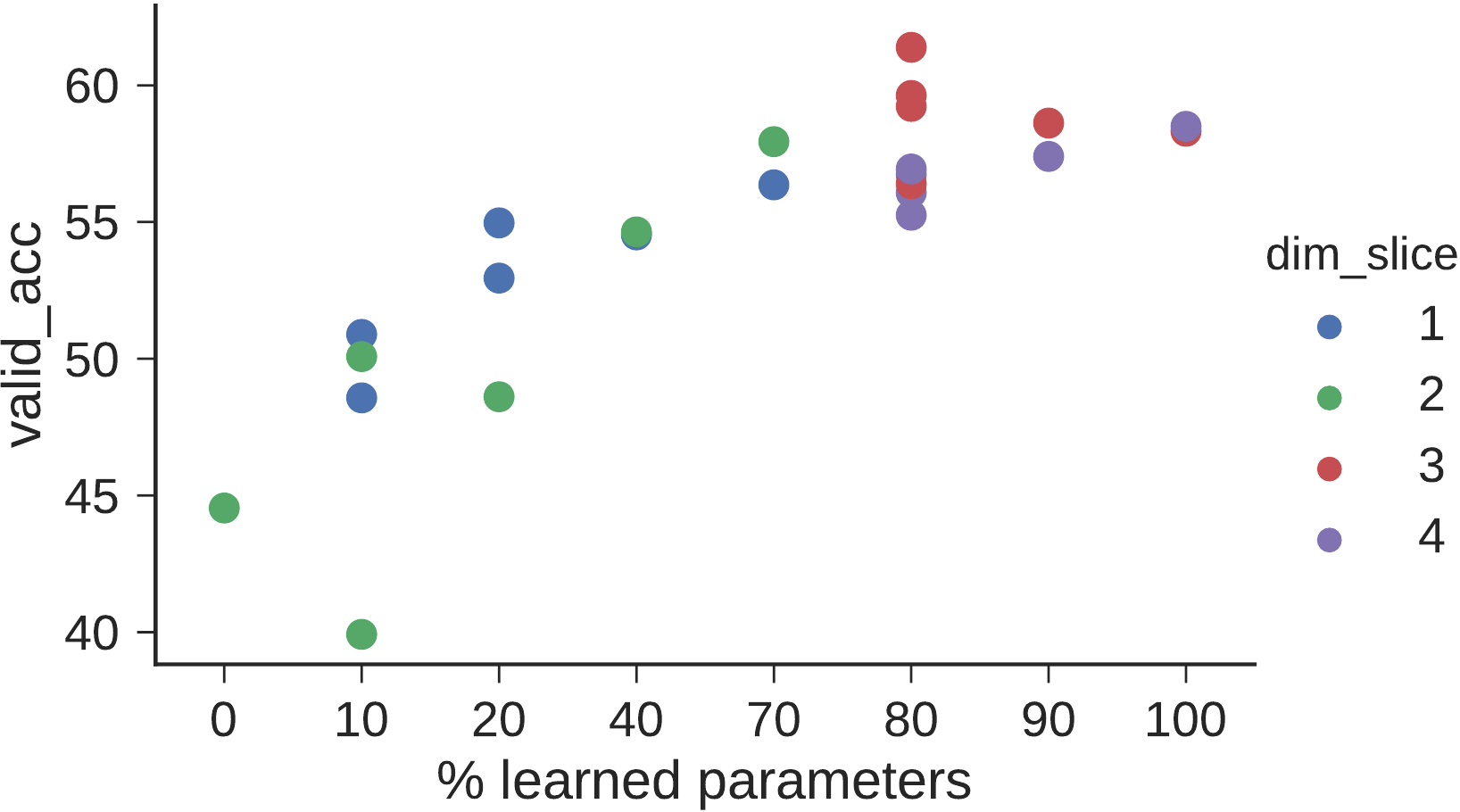}}\subfloat[]{\includegraphics[width=0.8\columnwidth]{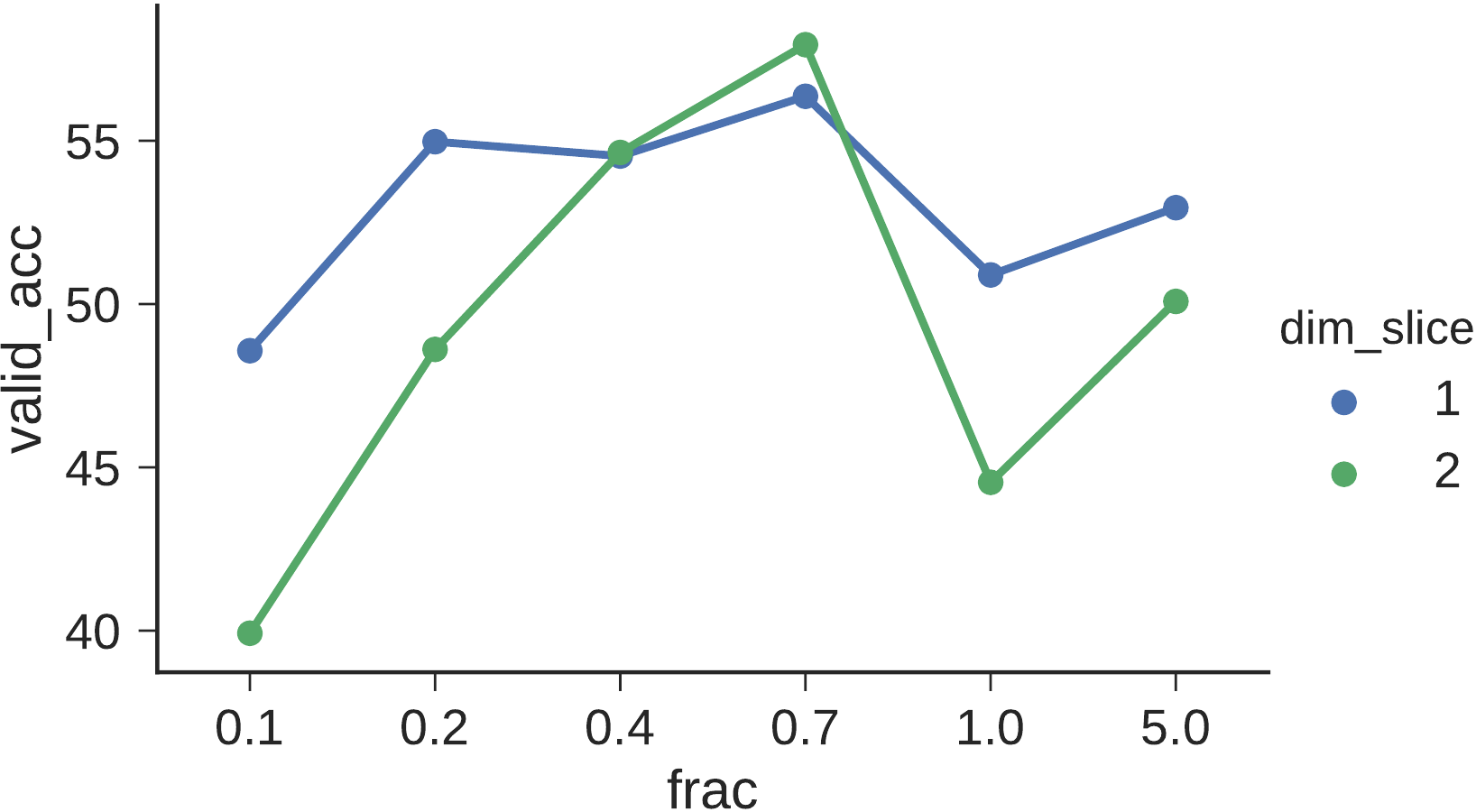}}
\par\end{centering}
\caption{\label{fig:various-dimensions}Training subsets of weights by slicing
different dimensions of the weight tensor (densenet, one epoch) (a)
fraction of total number of weights vs. selected dimension. dim\_slice
= 1: limiting output filters; 2: limiting input weights, 3,4: subsets
of weights according to spatial location inside kernel. (b) fraction
along selected dimension (\textless{}0) / size of slice along selected
dimension (\textgreater{}=1) . Learning a subset of filters outperforms
learning a subset of weights for each filter. }
\end{figure*}

\subsection{Splitting Network Parameters \label{subsec:Splitting-Network-Parameters}}

As we cannot test all possible subsets of network parameters, we define
a range of configurations to define some of them, as follows:

\textbf{Fractional Layers:} setting a constant fraction $p$ of filters
of each layer $L$ of $N$, except the fully-connected (classification)
layer. We do this for $p\in$\{$.07,.08,.09,.1,.4,.7\}$ (less than
0.07 will mean no filters for networks where $F_{0}=16$ \eg., in
WRN (wide resnets).

\textbf{Integer number of filters:} learning a constant integer $k\in\{1,5,10\}$
number of filters per layer. We show that learning a \emph{single
filter} \emph{in each layer }leads to non-trivial performance for
some architectures. 

\textbf{Single Layers:} freezing all weights except those of a single
block of layers. This is only for the wide and dense resnets. The
blocks are selected as $\mathcal{W}_{l}\in\{conv_{1}block_{1},block_{2},block_{3},fc\}$
where $conv_{1}$ is the first convolutional layer, $block_{i}$ is
one of 3 blocks of a wide-resnet with 28 layers and a widen factor
of 4 or of a densenet with a depth of 100 and a growth-rate of 12. 

\textbf{Batch Normalization: }we always use batch normalization layers
if they are part of the original model. However, the BN layers also
have optionally learnable parameters in addition to their running
estimates of mean and variance. These are a multiplication $\gamma$
and bias term $\beta$ for each filter. Hence, the number of total
weights in the BN layers amount to usually a few tens of thousands
- dependent on the number of convolutional layers followed by BN layers
and number of output channels for each. For example, in densenets
this equals $\sim24K$ parameters and $\sim18K$ in WRN networks with
a widen factor of 10. 

Although BN layers are usually thought of as auxiliary layers to improve
network convergence, we note that the benefits of learning the $\gamma,\beta$
parameters in a task dependent manner go beyond stabilization of the
optimization process. This is exemplified by the work of \cite{rebuffi2017learning}
where the parameters of the BN layers are trained for each task. In
our experiments, we take this a step further, to show what can be
learned by tuning \emph{only the BN parameters of the network}\textbf{\emph{.}}

\section{Experiments}

We experiment with the CIFAR-10 and CIFAR-100 datasets \cite{krizhevsky2009learning}
and several architectures: wide-residual networks, densely connected
resnets, AlexNet, and VGG-19 (resp. \cite{journals/corr/ZagoruykoK16,huang2016densely,krizhevsky2012imagenet,simonyan2014very},
as well as VGG-19 without batch-normalization (BN) \cite{ioffe2015batch}.
For baselines, we modify a reference implementation\footnote{\url{https://github.com/bearpaw/pytorch-classification}}.
To evaluate many different configurations, we limit the number of
training epochs for most of the experiments to 10 epochs (or even
one epoch for some cases). In these cases, we also reduce the widening
factor of the WRN to 4 from the default 10. For a few experiments
we perform a full run (200 or 300 epochs, depending on network architecture).

All experiments were performed using an Ubuntu machine with a single
Titan-X Pascal GPU, using the PyTorch\footnote{\url{http://pytorch.org/}}
deep-learning framework. Unless specified otherwise, models were optimized
using Momentum-SGD.

\subsection{Limiting Filter Inputs vs Outputs}

\begin{figure*}
\subfloat[]{\begin{centering}
\includegraphics[bb=0bp 0bp 888bp 337bp,clip,height=0.21\textwidth]{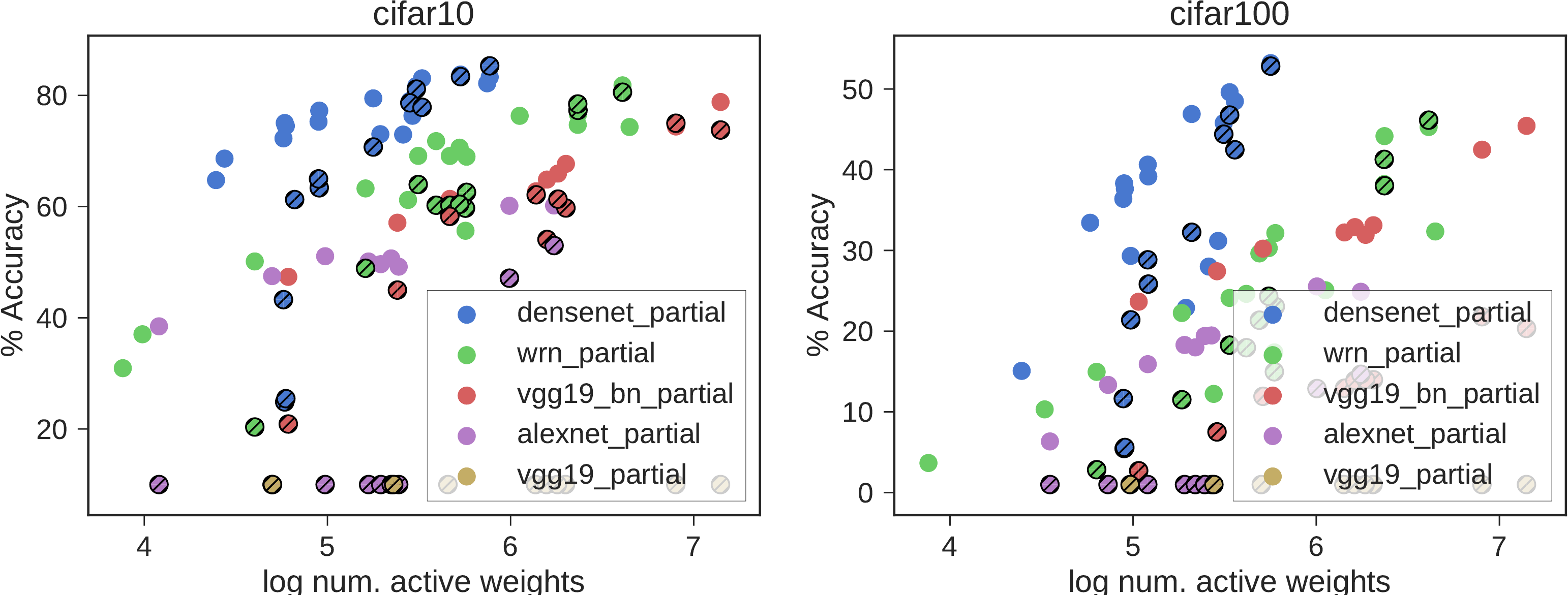}
\par\end{centering}
}\subfloat[]{\begin{centering}
\includegraphics[bb=0bp 0bp 888bp 337bp,clip,height=0.21\textwidth]{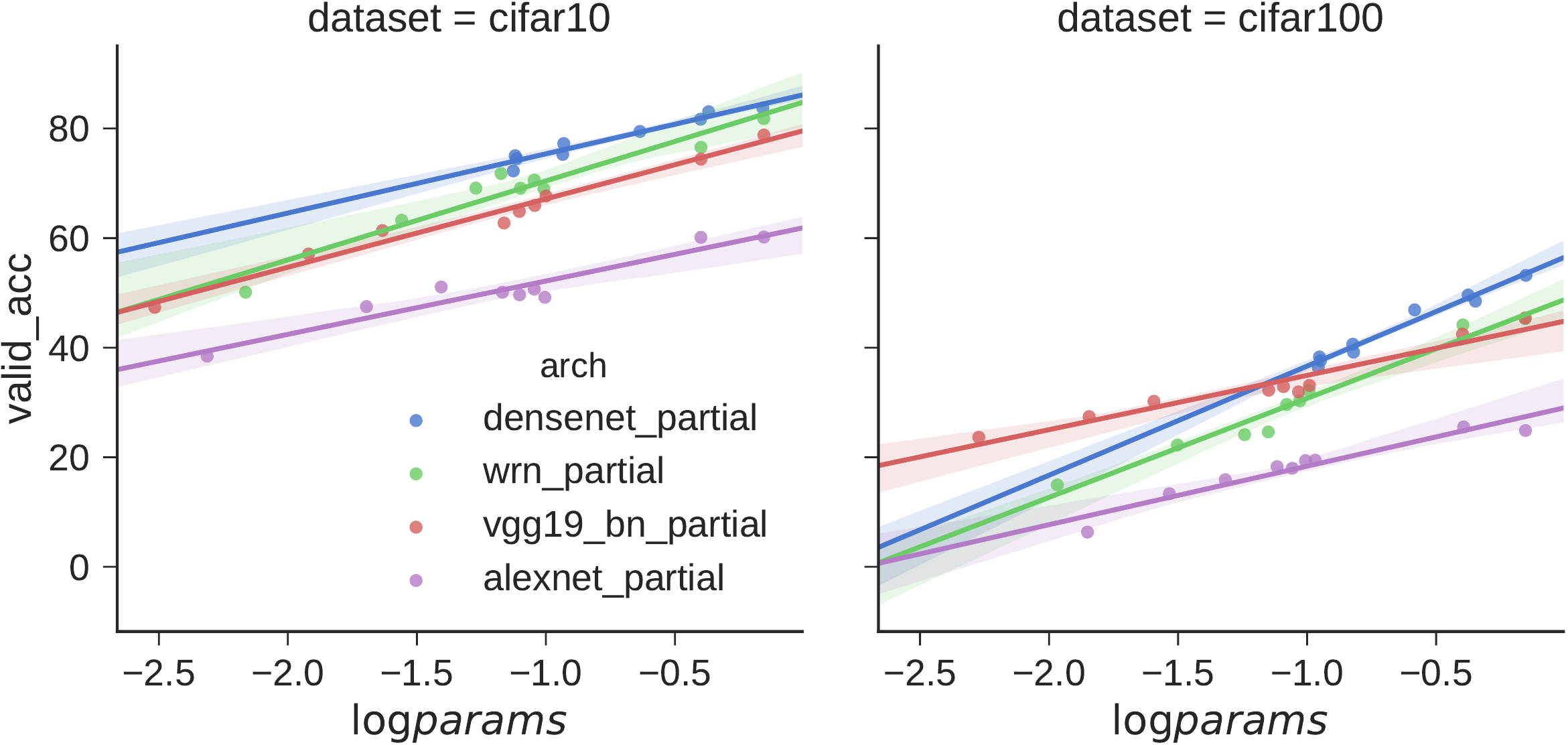}
\par\end{centering}
}
\centering{}\caption{\label{perf-vs-params}(a) Performance vs. absolute number of parameters
(log scale), after training for 10 epochs. Much of the performance
can be preserved by learning a relatively smaller no. of parameters,
and even zeroing out the rest (thatched circles). densenet (blue)
is very efficient in this sense, as it degrades most gradually with
the decrease in the number of learned parameters. (b) fitting lines
to performance vs. log fraction of learned parameters, this follows
a logarithmic curve for various models. }
\end{figure*}

As a first step, we establish in which dimension it is better to slice
the filter tensor $M_{i}$ in each layer. Recall (Section \ref{sec:Method})
that learning a subset of the first dimension of $M_{i}$ translates
to learning some of the filters but training all parameters for each
filter (\eg, considering all inputs), while learning the second to
fourth dimension means learning all filters but choosing some subset
of the filter coefficients for each. Intuitively, allowing a subset
of filters to be fully learned allows those learned filters to make
full-use of the features in their incoming connections. On the other
hand, learning all filters but selecting the \emph{same set of inputs
}for all of them to be fixed seems like a worse option, as this may
cause important features to be ``missed''; specifically, this can
happen due to fixed bias term which cause the non-linearities to zero
out the effect of these incoming features, rendering them ``invisible''
to subsequent layers. 

We have tested this with the densenet \cite{huang2016densely} architecture
on CIFAR-10, where we have sliced the convolutional layers along the
four different dimensions and limited the number of epochs to just
one. We tested this for a set fraction of each dimension, as well
as a fixed-sized slice from each dimension, with a minimum of one
element per slice. In Figure \ref{fig:various-dimensions} (a) we
plot the resulting performance vs the total number of parameters.
For each dimension we set the minimal size of the selected slice to
be 1. Because of this, slicing the third and fourth dimensions of
$M_{i}$ results in a much larger number of learned parameters. In
Figure \ref{fig:various-dimensions} (b) we show results of learning
slices of a given size along the first / second dimensions of $M_{i}$
for each layer. 

Indeed, learning only a subset of the filters (\emph{dim\_slice=1,
blue}) outperforms learning only a subset of the weights of each (\emph{dim\_slice=2,green}).
As a result, all experiments hereafter will only demonstrate configurations
involving subsets of filters (or entire layers). 

\subsection{Subsets of Filters}

With the conclusion from the section above, we turn to experiment
on various architectures and configurations, limiting ourselves to
selecting a subset of filters, using the ways described in Section
\ref{subsec:Splitting-Network-Parameters}. We discuss the results
below. 

Figure \ref{fig:Training-only-a} shows the top-1 accuracy after 10
training epochs. The best performer is densenets. AlexNet failed to
learn for non-trivial fractions or for only a few filters per layer.
Note the gap between the fixed weights and those zeroed-out. Zeroing
out the weights effectively reduces the number of filters from the
network. Using 70\% of filters while zeroing the rest out achieves
the same performance for densenets. 

Shaded areas in Figure \ref{fig:Training-only-a} specify learning
a constant, integer number of filters at each layer. Interestingly,
learning only a \emph{single filter per layer can }result in a non-trivial
accuracy. In fact, zeroing out all non-learned weight, resulting in
a net with a single-filter per layer, still is able to do much above
chance, around 45\% on CIFAR-10 with resnets. VGG-19 without the BN
layers failed to converge to performance better than chance for any
of these settings. 

\paragraph{Parameter Efficiency}

Another view on the results can be seen in Figure \ref{perf-vs-params}
(a) which plots the performance obtained vs. the total number of parameters
on a logarithmic scale. Even when zeroing out all non-learned parameters
(thatched circles), densenets attain decent performance with less
that 100K parameters, roughly an eighth of the original amount. We
fit a straight line to the performance vs. the log of the fraction
of learned parameters (Figure \ref{perf-vs-params} (b)). The performance
grows logarithmically with the number of parameters, with a larger
slope (\eg, better utilization of added parameters for recent architectures
(resnet, WRN). 

\paragraph{Subsets of Layers}

We learn only subsets of layers out of the layers specified and report
the performance for each of these scenarios, for CIFAR-10 and CIFAR-100
and for the dense and wide residual networks. Table \ref{tab:blocks}
summarizes this experiment. We see that a non-trivial accuracy can
be reached by learning only a single of the layer subsets. Furthermore,
in most cases learning the last, fully-connected layer on its own
proves inferior to doing so with another layer. For example, with
wide-resnets (WRN) learning the fc (fully connected) layer attains
only 37\% top-1 accuracy on CIFAR-10, much less than either than the
3 middle blocks. While the number of parameters in fc is indeed much
lower, note that it grows linearly with the number of classes while
that of the middle blocks remains constant (this is not seen directly
in the table due to the additional weights of BN layers learned).
From a practical point of view, this can indicate that when fine-tuning
a single layer of a network towards a new task, the last layer is
not necessarily the best one to choose. Nevertheless, fine-tuning
an additional layer in the middle can prove useful as the additional
parameter cost can be quite modest. 

\begin{table}
\begin{centering}
\small{\noindent\resizebox{\columnwidth}{!}{%
\begin{tabular}{l>{\raggedright}p{2cm}llll}
\toprule 
arch & Params & layer  & Eff. Params & C10 & C100\tabularnewline
\midrule
\multirow{6}{*}{densenet} & \multirow{6}{2cm}{0.77-\\
0.8M} & BN & 24K & 60.85 & 16.06\tabularnewline
 &  & $conv_{1}$ & 24.6K  & 64.76  & 15.08 \tabularnewline
 &  & $block_{1}$ & 194K  & 73.02  & 22.89 \tabularnewline
 &  & $block_{2}$ & 259K  & 72.95  & 28.00 \tabularnewline
 &  & $block_{3}$ & 291K  & 76.33  & 31.18 \tabularnewline
 &  & $fc$ & 27.4K / 58.3K & 68.63  & 33.43 \tabularnewline
\midrule
\multirow{6}{*}{WRN} & \multirow{6}{2cm}{5.85-\\
5.87M } & BN & 7.2K & 30.2 & 3.76\tabularnewline
 &  & $conv_{1}$ & 7.63K  & 30.93  & 3.67 \tabularnewline
 &  & $block_{1}$ & 275K  & 61.19  & 12.23 \tabularnewline
 &  & $block_{2}$ & 1.12M  & 76.32  & 25.07 \tabularnewline
 &  & $block_{3}$ & 4.46M  & 74.31  & 32.35 \tabularnewline
 &  & $fc$ & 9.77K / 32.9K & 37.03  & 10.31 \tabularnewline
\bottomrule
\end{tabular}}}
\par\end{centering}
\caption{\label{tab:blocks} Learning only a subset of layers (10 epochs):
learning only the fully-connected layer usually proves inferior to
learning one of the middle blocks (see $fc$ vs. $block_{i},i\in\{1,2,3\}$).
Learning \emph{only }the batch-norm (BN) layer results in non-trivial
performance. \emph{Param}s: total parameters. \emph{Eff. Param}s:
number of learned parameters. WRN: Wide-residual Networks.}
 
\end{table}

\paragraph{Batch-Norm Layers }

As mentioned in Section. \ref{subsec:Splitting-Network-Parameters}
we tested network performance when learning \emph{only }batch-normalization
layers. This experiment was done for the wide and dense residual network.
Learning \emph{only }the parameters of the BN layers can in non-trivial
performance using densenets, \eg 60.85\% for CIFAR-10 (vs 68.6 for
BN+fc) and 30.2. For CIFAR-100 this is no longer the case, 16\% vs
33.4\%. Please refer to Table \ref{tab:blocks}. 

\begin{figure*}
\begin{centering}
\includegraphics[width=0.5\textwidth]{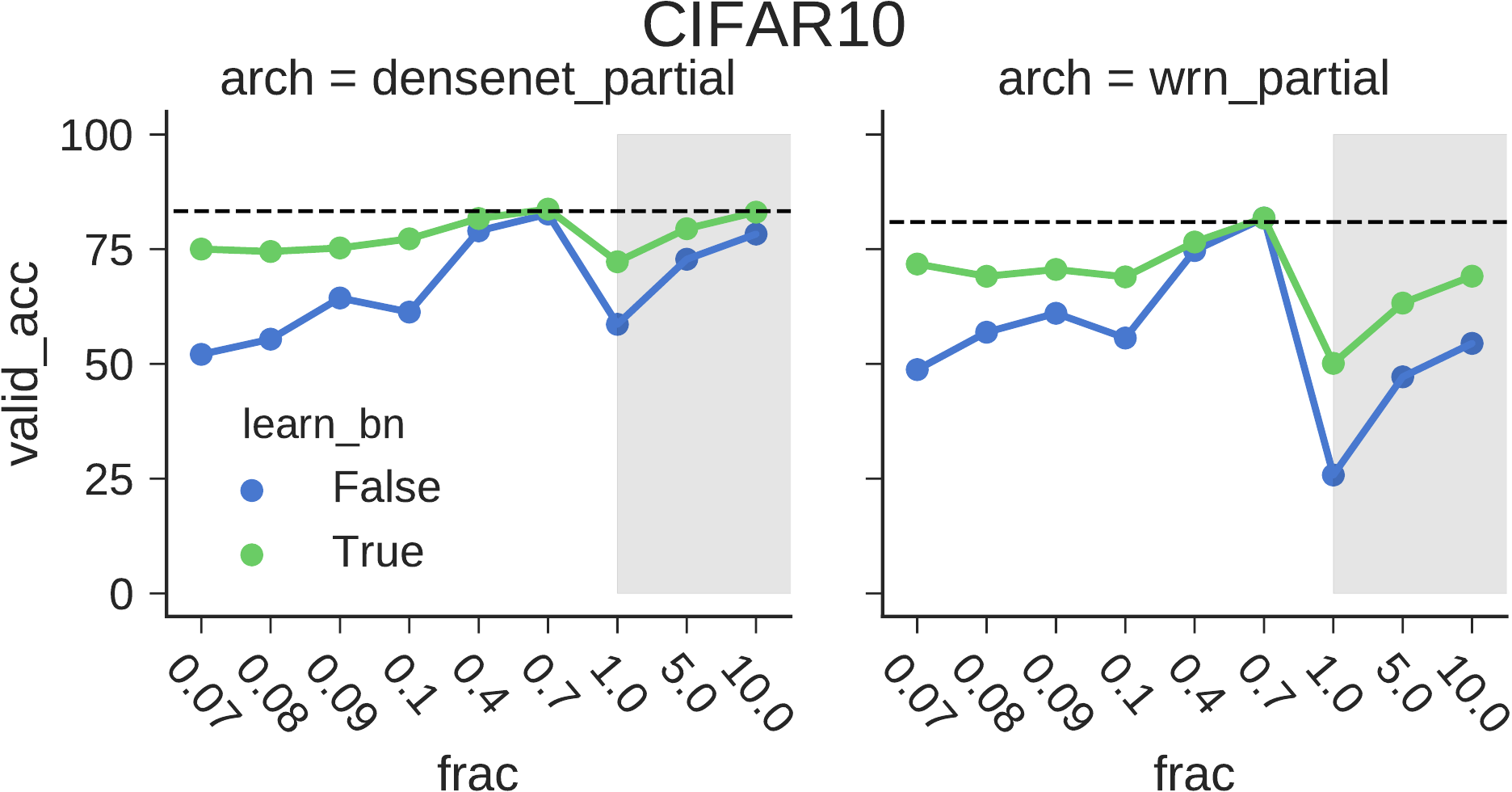}\includegraphics[width=0.5\textwidth]{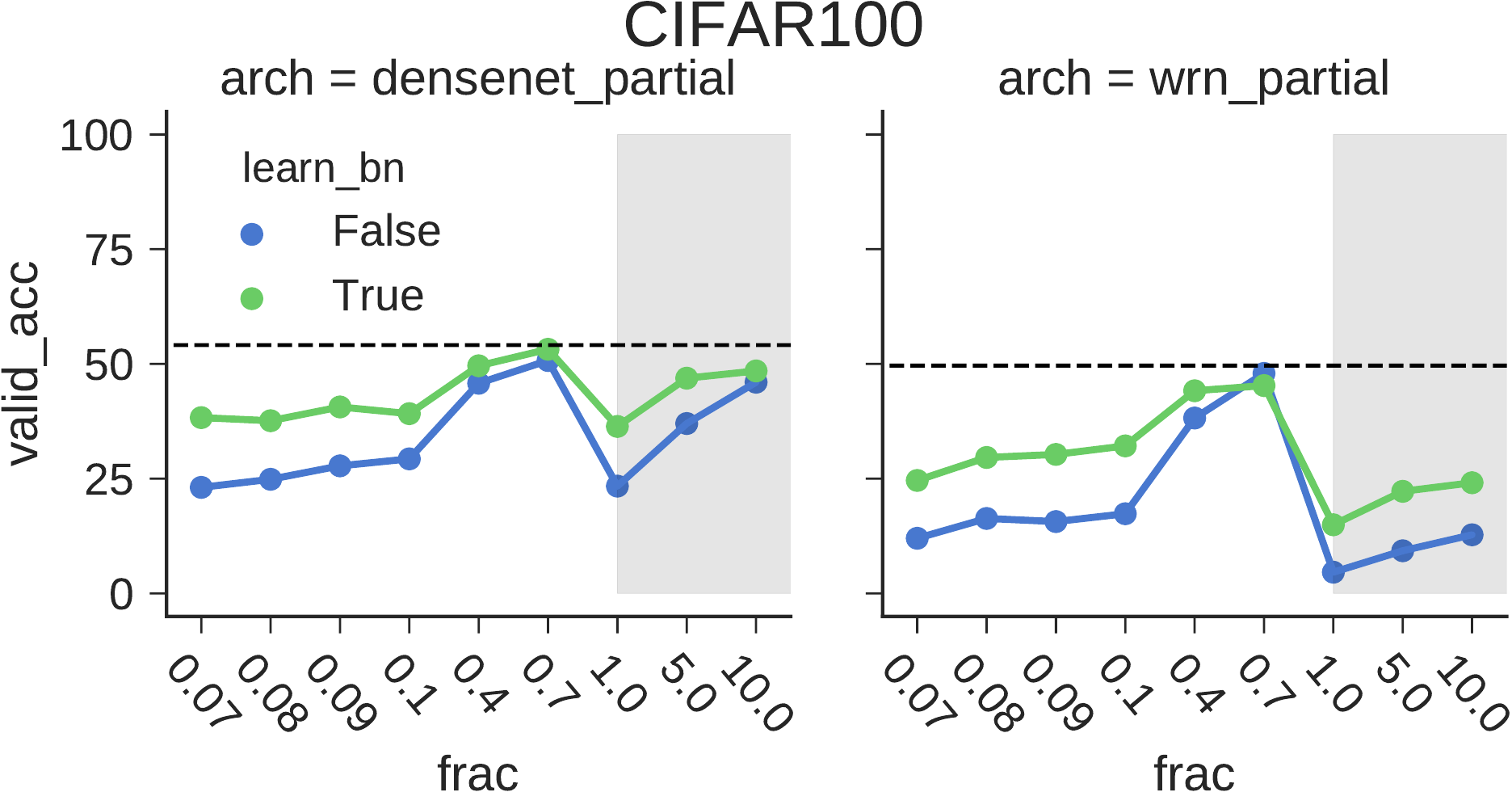}
\par\end{centering}
\caption{\label{fig:batch-norm}Learning different fractions of a network (10
epochs), with and without batch-norm layers. Batch-normalization accounts
for much of the accuracy when the number of weights is reduced but
become less significant when using a large (\textgreater{}40\%) fraction
of weights. Black dashed lines are for using all-weights with batch-norm. }
\end{figure*}

In addition, we compare the performance for various learned network
fractions when batch-norm is turned on and off for WRN and densenets.
This is summarized in Figure \ref{fig:batch-norm}. We can see that
when a small fraction of parameters ($\leq0.1$, or a constant integer
number of filters per layer) is used, BN layers make a big difference.
However, starting from $0.4$ the difference becomes smaller. This
is likely because the representative power introduced by the BN parameters
becomes less significant as more parameters are introduced (\ie,
learned by the optimizer). 

In this figure, we also see the performance attainable by training
100\% of the weights for each architecture (black dashed lines). Notably,
using 70\% percent of parameters induces little or no loss in accuracy
- and for densenet, we can achieve the full-accuracy with 10 filters
per layer. From this we see that there are various ways to distribute
the fraction of parameters trained in the network to achieve similar
accuracies. 

Achieving high accuracy with as much as 40\% filters learned is also
consistent with the result we got on the full training runs (see below).

\paragraph{Full Runs}

We also ran full training sessions for WRN and densenet (200,300 epochs
respectively) with a limited no. of parameters on CIFAR-10. Specifically,
WRN , when 60\% of the filters are arbitrarily zeroed out, achieve
almost the baseline performance of 96.2\%. Please refer to Table \ref{tab:(a)-full-runs}
for numerical results of this experiment. We note that when learning
a single filter for each layer, densenets outperform WRN by a large
gap, 85\% vs 69\% on CIFAR-10 and 59.7\% vs 34.85 on CIFAR-100. 

\begin{figure}
\includegraphics[width=1\columnwidth]{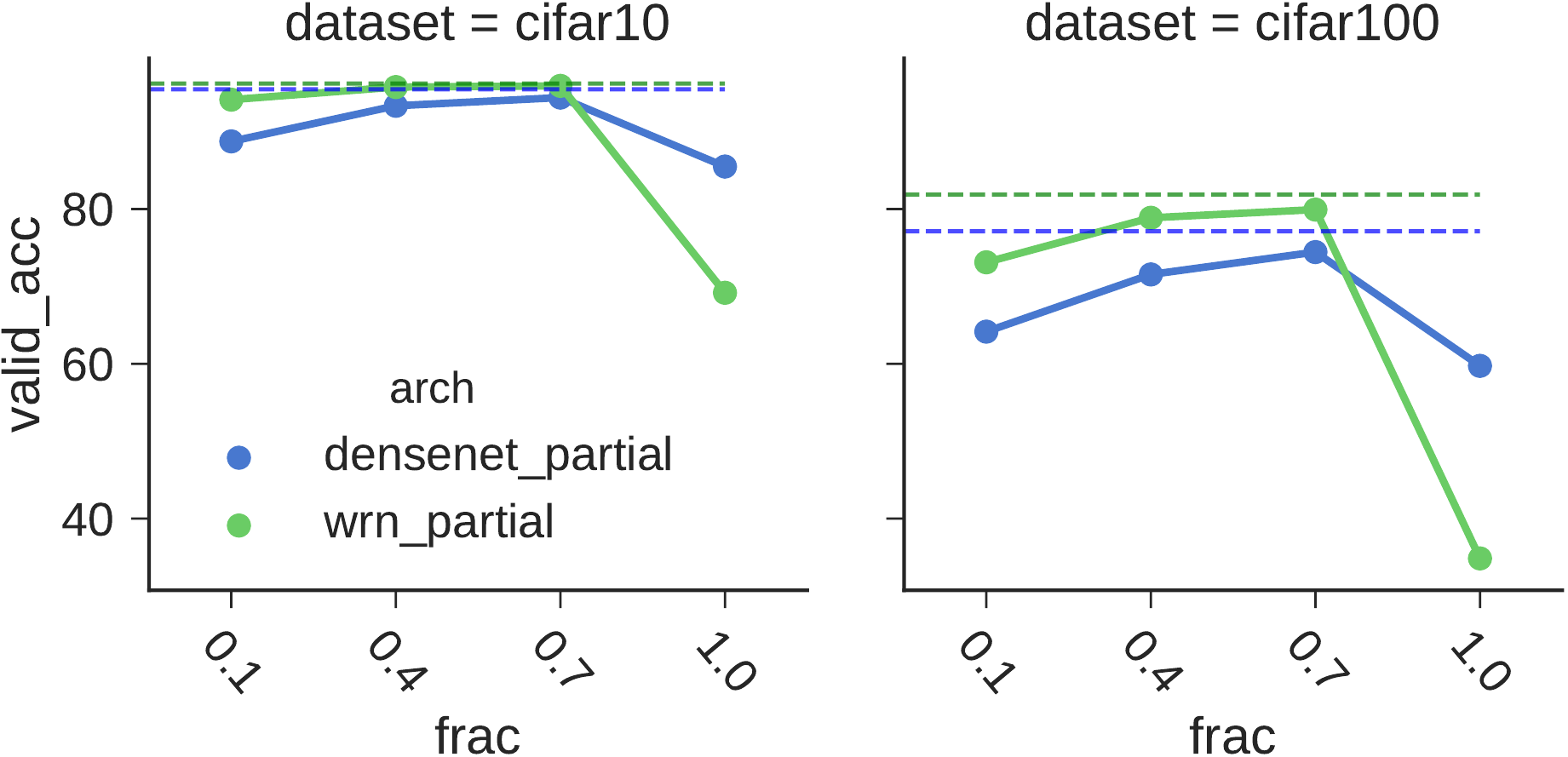}

\caption{\label{fig:Learning-full}Learning only a fraction of weights, full
runs: learning CIFAR-10 can be done effectively when most of the weights
are fixed at their random initialized state(\emph{left}). For CIFAR-100
a larger portion of parameters is needed (\emph{right}). Green and
blue dashed lines represent results with all weight learned for WRN,
densenets respectively. }
\end{figure}

On CIFAR-100, the gap between the accuracy attained with a fraction
of parameters vs all of them is larger when using a few parameter
than on CIFAR-10. This is visualized in Figure \ref{fig:Learning-full}. 
\begin{center}
{\small{}}
\begin{table}
\begin{centering}
\small{{\small{}}%
\begin{tabular}{ccccc}
\toprule 
Method & Fraction & Eff. Params $\times10^{6}$ & Perf & Perf$\dagger$\tabularnewline
\midrule 
WRN & 0.1 & 3.66 & 94.12  & 91.53\tabularnewline
\midrule 
WRN & 0.4 & 14.6 & 95.75 & 95.49\tabularnewline
\midrule 
densenets & 0.1 & 0.09 & 88.73 & 82.11\tabularnewline
\midrule 
densenets & 0.4 & 0.3 & 93.33  & 92.46\tabularnewline
\bottomrule
\end{tabular}{\small{}}}
\par\end{centering}{\small \par}
{\small{}\caption{\label{tab:(a)-full-runs}Performance vs fraction of parameters learned
on CIFAR-10 (full training, 200/300 epochs). $\dagger$ means performance
when $w_{f}$ (fixed parameters) are \emph{all set to zero}. Eff.
Params means number of params updated in learning.}
}{\small \par}
\end{table}
\par\end{center}{\small \par}

\paragraph{Tiny-ImageNet \label{par:Tiny-ImageNet}}

\begin{table}
\begin{centering}
\small{%
\begin{tabular}{ccc}
\toprule 
\% Params & Top-1 Accuracy \% & \# Params\tabularnewline
\midrule 
10 & 21.75 & .83M\tabularnewline
\midrule 
40 & 30.13 & 2.58M\tabularnewline
\midrule 
70 & 33.22 & 4.33M\tabularnewline
\midrule 
100 & 35.54 & 6.1M\tabularnewline
\bottomrule
\end{tabular}}
\par\end{centering}
\caption{\label{tab:tiny-imagenet}Training partial models on the Tiny-ImageNet
dataset.}
\end{table}

Finally, we perform a larger scale experiment. We choose the Tiny-ImageNet
\footnote{\url{https://tiny-imagenet.herokuapp.com/}}. This dataset
is a variant of the larger ImageNet \cite{russakovsky2015imagenet}
dataset. It contains two-hundred image categories, with 500 training
and 50 validation images for each. The images are downscaled to 64x64
pixels. In this experiment we use the recent YellowFin optimizer \cite{1706.03471v1}
for which we found there is less need for manual tuning than SGD.
We train for a total 45 epochs, with an initial learning rate of 0.1,
which is lowered by a factor of 10 after 15 and after 30 iterations.
We use the WRN architecture with a widen-factor of 4. The same architecture
was shown to perform quite reasonably well on another downs-scaled
version of ImageNet in \cite{rebuffi2017learning}.  We train the
fully-parameterized version, and partial versions with fractions of
0.1,0.4 and 0.7 of the filters in each conv. layer. 

The results of this experiment are summarized in Table \ref{tab:tiny-imagenet}.
The value of the fully-parameterized version (last row in the table)
is of little importance. Note how with 70\% of the parameters trained,
we lose \%2.3 in the top-1 accuracy, and another 3.1\% for 40\% of
the convolutional layers. 

\subsection{Cheap Ensembles}

Training a network while keeping most weights fixed enables the creation
of ``cheap'' ensemble models that share many weights and vary only
the remaining portion. For example, training a densenet model while
learning only 10\% of the weights requires roughly 90K new parameters
for each such model. The total cost for \eg, an ensemble model of
size 5 will be $.77+09*4=1.13M$ parameters, much less than training
five independent models. But will the resulting ensemble be as diverse
as five independently trained models? Using densenets and testing
on CIFAR-10, we trained three ensemble models of 5 elements each.
We first train a fully-parametrized model for one epoch and use it
as a starting point to train each ensemble element for an additional
epoch. In this experiment, we tried both the Adam \cite{kingma2014adam}
Optimizer and SGD and reported for each ensemble type the best of
the two results. 

We report for each ensemble the mean accuracy of elements, the accuracy
attained averaging the ensemble's predictions and the total number
of learned parameters (except of the first fully-parametrized model).
We trained ensembles of (a) fully parameterized models (\emph{Full})
(b) models varying only by the fc layer (\emph{FC}) and (c) models
with a shared fraction $R$ of convolutional weights (\emph{Share-Conv-R).}
Table \ref{tab:Ensemble-models.-An} summarizes the results. We see
that fixing 10\% (sharing 90\%) of the parameters already outperforms
re-training only the FC layers. A fully-parameterized ensemble indeed
shows better variability in the solutions (leading to a better ensemble
performance), though using some portion of shared conv. weights is
not far below it, with significantly less weights. 

\begin{table}
\small{\noindent\resizebox{\columnwidth}{!}{%
\begin{tabular}{cccc}
\toprule 
Model Type & No. Params & Mean Accuracy & Ensemble Accuracy\tabularnewline
\midrule 
FC & 137K & 68.5 & 70.25\tabularnewline
\midrule 
Share-Conv-0.9 & 451K & 69.9 & 71.14\tabularnewline
\midrule 
Share-Conv-0.6 & 1.53M & 71 & 73.54\tabularnewline
\midrule 
Share-Conv-0.3 & 2.66M & 71.3 & 75.47\tabularnewline
\midrule 
Full & 3.85M & 71.7 & 76.35\tabularnewline
\bottomrule
\end{tabular}}}

\caption{Ensemble models. \label{tab:Ensemble-models.-An}Ensemble models with
some convolutional weight sharing between elements perform similarly
to independently trained models though being significantly more compact.
Relearning some of the conv. weights }
\end{table}

\subsection{Weight Magnitudes}

We perform an analysis of the magnitude distribution of the weights
within learned vs. fixed layers. This is motivated by the observation
that relatively small weights can lead to better generalization bounds
\cite{bartlett1997valid,zhou2017landscape}. We analyze the magnitudes
of the weights of the convolutional layers of the experiment in the
above Section \ref{par:Tiny-ImageNet} (on the Tiny-ImageNet dataset).
For each convolutional layer, we record the mean of the absolute weight
value and the variance of the weights. For layers with fixed weights,
we report this value as well. The mean and variance of the fixed weights
is determined by the random initialization. 

\begin{figure}
\includegraphics[width=1\columnwidth]{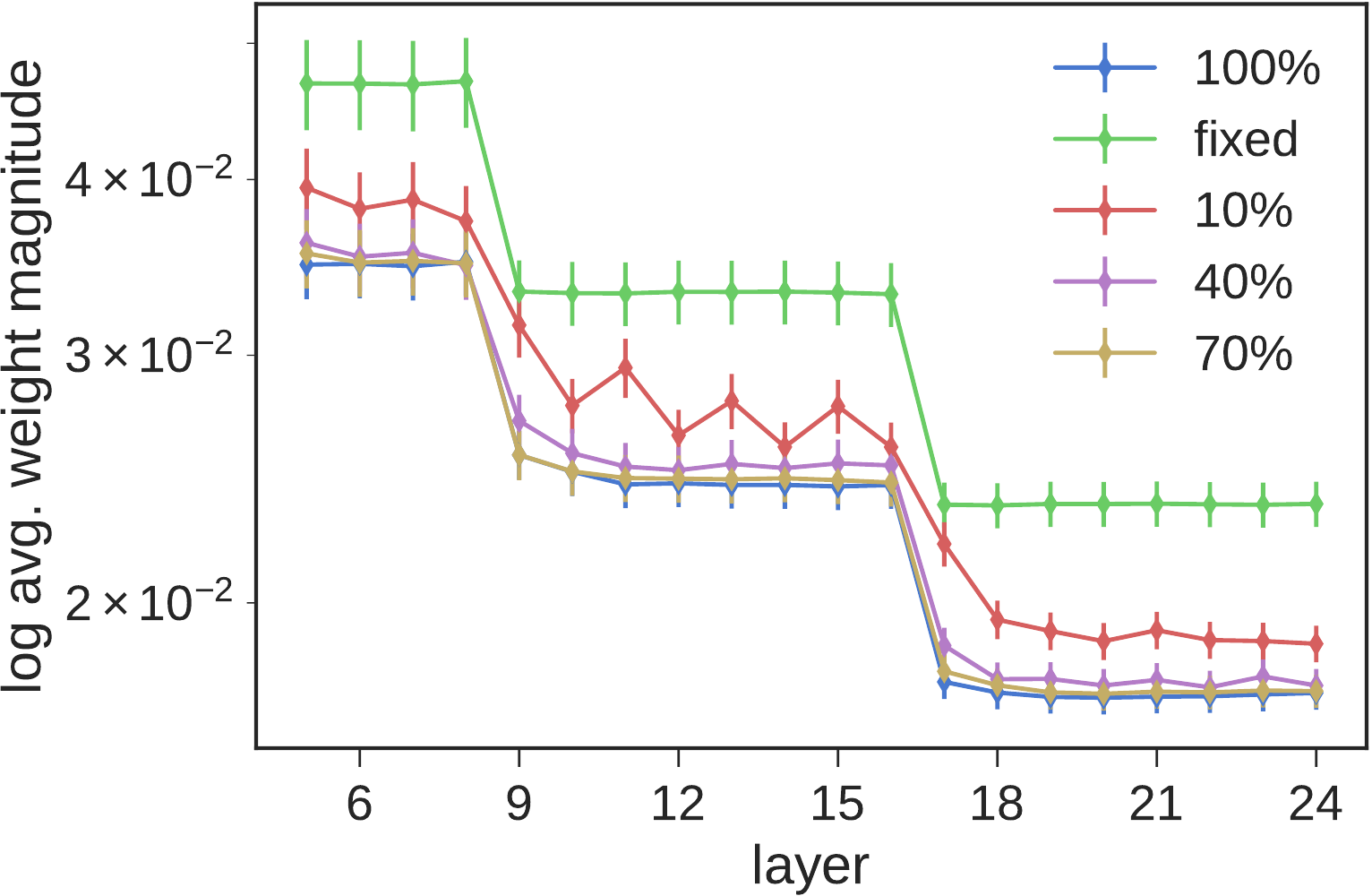}

\caption{\label{fig:Weight-distribution-across}Weight distribution across
convolutional layers as a function of the percentage of learned convolutional
weights. The mean value of any fraction above 40\% of trained parameters
is stable and similar to that of a fully-trained network. ``Fixed''
is simply the distribution of randomly-initialized weights. }
\end{figure}

We plot the results in Figure \ref{fig:Weight-distribution-across}.
The magnitudes of training only 10\% of the convolutional weights
stand out as being relatively high. For training 40\%, 70\% and 100\%
(all weights), the magnitudes seem to be distributed quite similarly,
with 70\% and 100\% being nearly indistinguishable. If the average
magnitude of weights is indeed any indication of the generalization
capacity of the network, in this context it seems to be consistent
with the rest of our findings, specifically the performance of the
partially trained nets as seen in Table \ref{tab:tiny-imagenet}.

\section{Discussion}

We have demonstrated that learning only a small subset of the parameters
of the network or a subset of the layers leads to an unexpectedly
small decrease in performance (with respect to full learning) - even
though the remaining parameters are either fixed or zeroed out. This
is contrary to common practice of training all network weights. We
hypothesize this shows how over-parameterized current models are,
even those with a relatively small number of parameters, such as densenets.
We have tested a large number of configurations of ways to limit the
subsets of learned network weights. This choice of subset has a large
impact on the performance of the resulting network. Some network architectures
are more robust to fixing most of their weights than others. Learning
is also possible with an extremely small number of filters learned
at the convolutional layers, as little as a single filter for each
of these layers. Three simple applications of the described phenomena
are (1) cheap ensemble models, all with the same ``backbone'' fixed
network, (2) learning multiple representations with a small number
of parameters added to each new task and (3) transfer-learning (or
simply learning) by learning a \emph{middle }layer vs the final classification
layer. We intend to explore these directions in future work, as well
as testing the reported phenomena on additional, non-vision related
tasks, such as natural language processing or reinforcement-learning. 

\bibliographystyle{IEEEtran}
\bibliography{example_paper_w_ensemble_arxiv}

\begin{thebibliography}{10}
\providecommand{\url}[1]{#1}
\csname url@samestyle\endcsname
\providecommand{\newblock}{\relax}
\providecommand{\bibinfo}[2]{#2}
\providecommand{\BIBentrySTDinterwordspacing}{\spaceskip=0pt\relax}
\providecommand{\BIBentryALTinterwordstretchfactor}{4}
\providecommand{\BIBentryALTinterwordspacing}{\spaceskip=\fontdimen2\font plus
\BIBentryALTinterwordstretchfactor\fontdimen3\font minus
  \fontdimen4\font\relax}
\providecommand{\BIBforeignlanguage}[2]{{%
\expandafter\ifx\csname l@#1\endcsname\relax
\typeout{** WARNING: IEEEtran.bst: No hyphenation pattern has been}%
\typeout{** loaded for the language `#1'. Using the pattern for}%
\typeout{** the default language instead.}%
\else
\language=\csname l@#1\endcsname
\fi
#2}}
\providecommand{\BIBdecl}{\relax}
\BIBdecl

\bibitem{raghu2016expressive}
M.~Raghu, B.~Poole, J.~Kleinberg, S.~Ganguli, and J.~Sohl-Dickstein, ``{On the
  expressive power of deep neural networks},'' \emph{arXiv preprint
  arXiv:1606.05336}, 2016.

\bibitem{shwartz2017opening}
R.~Shwartz-Ziv and N.~Tishby, ``{Opening the Black Box of Deep Neural Networks
  via Information},'' \emph{arXiv preprint arXiv:1703.00810}, 2017.

\bibitem{he2016identity}
K.~He, X.~Zhang, S.~Ren, and J.~Sun, ``{Identity mappings in deep residual
  networks},'' in \emph{{European Conference on Computer Vision}}.\hskip 1em
  plus 0.5em minus 0.4em\relax Springer, 2016, pp. 630--645.

\bibitem{krizhevsky2012imagenet}
A.~Krizhevsky, I.~Sutskever, and G.~E. Hinton, ``{Imagenet classification with
  deep convolutional neural networks},'' in \emph{{Advances in neural
  information processing systems}}, 2012, pp. 1097--1105.

\bibitem{simonyan2014very}
K.~Simonyan and A.~Zisserman, ``{Very deep convolutional networks for
  large-scale image recognition},'' \emph{arXiv preprint arXiv:1409.1556},
  2014.

\bibitem{journals/corr/ZagoruykoK16}
S.~Zagoruyko and N.~Komodakis, ``{Wide Residual Networks.}'' \emph{CoRR}, vol.
  abs/1605.07146, 2016.

\bibitem{shazeer2017outrageously}
N.~Shazeer, A.~Mirhoseini, K.~Maziarz, A.~Davis, Q.~Le, G.~Hinton, and J.~Dean,
  ``{Outrageously large neural networks: The sparsely-gated mixture-of-experts
  layer},'' \emph{arXiv preprint arXiv:1701.06538}, 2017.

\bibitem{krizhevsky2009learning}
A.~Krizhevsky and G.~Hinton, ``{Learning multiple layers of features from tiny
  images},'' 2009.

\bibitem{russakovsky2015imagenet}
O.~Russakovsky, J.~Deng, H.~Su, J.~Krause, S.~Satheesh, S.~Ma, Z.~Huang,
  A.~Karpathy, A.~Khosla, M.~Bernstein \emph{et~al.}, ``{Imagenet large scale
  visual recognition challenge},'' \emph{International Journal of Computer
  Vision}, vol. 115, no.~3, pp. 211--252, 2015.

\bibitem{li2016pruning}
H.~Li, A.~Kadav, I.~Durdanovic, H.~Samet, and H.~P. Graf, ``{Pruning filters
  for efficient convnets},'' \emph{arXiv preprint arXiv:1608.08710}, 2016.

\bibitem{1801.10447}
D.~Mittal, S.~Bhardwaj, M.~M. Khapra, and B.~Ravindran, ``{Recovering from
  Random Pruning: On the Plasticity of Deep Convolutional Neural Networks},''
  2018.

\bibitem{huang2015trends}
G.~Huang, G.-B. Huang, S.~Song, and K.~You, ``{Trends in extreme learning
  machines: A review},'' \emph{Neural Networks}, vol.~61, pp. 32--48, 2015.

\bibitem{rudi2017generalization}
A.~Rudi and L.~Rosasco, ``{Generalization properties of learning with random
  features},'' in \emph{{Advances in Neural Information Processing Systems}},
  2017, pp. 3218--3228.

\bibitem{giryes2015deep}
R.~Giryes, G.~Sapiro, and A.~M. Bronstein, ``{Deep Neural Networks with Random
  Gaussian Weights: A Universal Classification Strategy?}'' \emph{arXiv
  preprint arXiv:1504.08291}, 2015.

\bibitem{hoffer2018fix}
E.~Hoffer, I.~Hubara, and D.~Soudry, ``{Fix your classifier: the marginal value
  of training the last weight layer},'' \emph{arXiv preprint arXiv:1801.04540},
  2018.

\bibitem{horadam2012hadamard}
K.~J. Horadam, \emph{{Hadamard matrices and their applications}}.\hskip 1em
  plus 0.5em minus 0.4em\relax Princeton university press, 2012.

\bibitem{han2015deep}
S.~Han, H.~Mao, and W.~J. Dally, ``{Deep compression: Compressing deep neural
  networks with pruning, trained quantization and huffman coding},''
  \emph{arXiv preprint arXiv:1510.00149}, 2015.

\bibitem{han2015learning}
S.~Han, J.~Pool, J.~Tran, and W.~Dally, ``{Learning both weights and
  connections for efficient neural network},'' in \emph{{Advances in Neural
  Information Processing Systems}}, 2015, pp. 1135--1143.

\bibitem{liu2015sparse}
B.~Liu, M.~Wang, H.~Foroosh, M.~Tappen, and M.~Pensky, ``{Sparse convolutional
  neural networks},'' in \emph{{Proceedings of the IEEE Conference on Computer
  Vision and Pattern Recognition}}, 2015, pp. 806--814.

\bibitem{wen2016learning}
W.~Wen, C.~Wu, Y.~Wang, Y.~Chen, and H.~Li, ``{Learning structured sparsity in
  deep neural networks},'' in \emph{{Advances in Neural Information Processing
  Systems}}, 2016, pp. 2074--2082.

\bibitem{iandola2016squeezenet}
F.~N. Iandola, S.~Han, M.~W. Moskewicz, K.~Ashraf, W.~J. Dally, and K.~Keutzer,
  ``{SqueezeNet: AlexNet-level accuracy with 50x fewer parameters and< 0.5 MB
  model size},'' \emph{arXiv preprint arXiv:1602.07360}, 2016.

\bibitem{howard2017mobilenets}
A.~G. Howard, M.~Zhu, B.~Chen, D.~Kalenichenko, W.~Wang, T.~Weyand,
  M.~Andreetto, and H.~Adam, ``{Mobilenets: Efficient convolutional neural
  networks for mobile vision applications},'' \emph{arXiv preprint
  arXiv:1704.04861}, 2017.

\bibitem{shayar2017learning}
O.~Shayar, D.~Levi, and E.~Fetaya, ``{Learning Discrete Weights Using the Local
  Reparameterization Trick},'' \emph{arXiv preprint arXiv:1710.07739}, 2017.

\bibitem{rastegari2016xnor}
M.~Rastegari, V.~Ordonez, J.~Redmon, and A.~Farhadi, ``{Xnor-net: Imagenet
  classification using binary convolutional neural networks},'' in
  \emph{{European Conference on Computer Vision}}.\hskip 1em plus 0.5em minus
  0.4em\relax Springer, 2016, pp. 525--542.

\bibitem{zeiler2014visualizing}
M.~D. Zeiler and R.~Fergus, ``{Visualizing and understanding convolutional
  networks},'' in \emph{{European conference on computer vision}}.\hskip 1em
  plus 0.5em minus 0.4em\relax Springer, 2014, pp. 818--833.

\bibitem{zhou2016learning}
B.~Zhou, A.~Khosla, A.~Lapedriza, A.~Oliva, and A.~Torralba, ``{Learning deep
  features for discriminative localization},'' in \emph{{Computer Vision and
  Pattern Recognition (CVPR), 2016 IEEE Conference on}}.\hskip 1em plus 0.5em
  minus 0.4em\relax IEEE, 2016, pp. 2921--2929.

\bibitem{biparva2017stnet}
M.~Biparva and J.~Tsotsos, ``{STNet: Selective Tuning of Convolutional Networks
  for Object Localization},'' in \emph{{The IEEE International Conference on
  Computer Vision (ICCV)}}, vol.~2, 2017.

\bibitem{zhang2016top}
J.~Zhang, Z.~Lin, J.~Brandt, X.~Shen, and S.~Sclaroff, ``{Top-down neural
  attention by excitation backprop},'' in \emph{{European Conference on
  Computer Vision}}.\hskip 1em plus 0.5em minus 0.4em\relax Springer, 2016, pp.
  543--559.

\bibitem{simonyan2013deep}
K.~Simonyan, A.~Vedaldi, and A.~Zisserman, ``{Deep inside convolutional
  networks: Visualising image classification models and saliency maps},''
  \emph{arXiv preprint arXiv:1312.6034}, 2013.

\bibitem{mahendran2015understanding}
A.~Mahendran and A.~Vedaldi, ``{Understanding deep image representations by
  inverting them},'' in \emph{{Computer Vision and Pattern Recognition (CVPR),
  2015 IEEE Conference on}}.\hskip 1em plus 0.5em minus 0.4em\relax IEEE, 2015,
  pp. 5188--5196.

\bibitem{bau2017network}
D.~Bau, B.~Zhou, A.~Khosla, A.~Oliva, and A.~Torralba, ``{Network Dissection:
  Quantifying Interpretability of Deep Visual Representations},'' \emph{arXiv
  preprint arXiv:1704.05796}, 2017.

\bibitem{dong2017towards}
Y.~Dong, H.~Su, J.~Zhu, and F.~Bao, ``{Towards interpretable deep neural
  networks by leveraging adversarial examples},'' \emph{arXiv preprint
  arXiv:1708.05493}, 2017.

\bibitem{fong2018net2vec}
R.~Fong and A.~Vedaldi, ``{Net2Vec: Quantifying and Explaining how Concepts are
  Encoded by Filters in Deep Neural Networks},'' \emph{arXiv preprint
  arXiv:1801.03454}, 2018.

\bibitem{rebuffi2017learning}
S.-A. Rebuffi, H.~Bilen, and A.~Vedaldi, ``{Learning multiple visual domains
  with residual adapters},'' \emph{arXiv preprint arXiv:1705.08045}, 2017.

\bibitem{huang2016densely}
G.~Huang, Z.~Liu, K.~Q. Weinberger, and L.~van~der Maaten, ``{Densely connected
  convolutional networks},'' \emph{arXiv preprint arXiv:1608.06993}, 2016.

\bibitem{ioffe2015batch}
S.~Ioffe and C.~Szegedy, ``{Batch normalization: Accelerating deep network
  training by reducing internal covariate shift},'' in \emph{{International
  Conference on Machine Learning}}, 2015, pp. 448--456.

\bibitem{1706.03471v1}
J.~Zhang, I.~Mitliagkas, and C.~R{\'e}, ``{YellowFin and the Art of Momentum
  Tuning},'' \emph{arXiv preprint arXiv:1706.03471}, 2017.

\bibitem{kingma2014adam}
D.~Kingma and J.~Ba, ``{Adam: A method for stochastic optimization},''
  \emph{arXiv preprint arXiv:1412.6980}, 2014.

\bibitem{bartlett1997valid}
P.~L. Bartlett, ``{For valid generalization the size of the weights is more
  important than the size of the network},'' in \emph{{Advances in neural
  information processing systems}}, 1997, pp. 134--140.

\bibitem{zhou2017landscape}
P.~Zhou and J.~Feng, ``{The Landscape of Deep Learning Algorithms},''
  \emph{arXiv preprint arXiv:1705.07038}, 2017.

\end{thebibliography}

\end{document}